\documentclass[sigconf]{acmart}

\AtBeginDocument{%
  }

\setcopyright{acmlicensed}
\copyrightyear{2018}
\acmYear{2018}
\acmDOI{XXXXXXX.XXXXXXX}

\acmConference[Conference acronym 'XX]{Make sure to enter the correct
  conference title from your rights confirmation email}{June 03--05,
  2018}{Woodstock, NY}

\acmISBN{978-1-4503-XXXX-X/2018/06}
\usepackage{multirow}
\usepackage{algorithm}
\usepackage{algorithmic}
\usepackage{subcaption}

\begin{document}
\begin{sloppypar}

\title{Causality-aligned Prompt Learning via Diffusion-based Counterfactual Generation}

\author{Xinshu Li}
\email{xinshu.li@unsw.edu.au}
\affiliation{%
  \institution{The University of New South Wales}
  \city{Sydney}
  \country{Australia}
}
\author{Ruoyu Wang}
\email{ruoyu.wang5@unsw.edu.au}
\affiliation{%
  \institution{The University of New South Wales}
  \city{Sydney}
  \country{Australia}
}
\author{Erdun Gao}
\email{erdun.gao@adelaide.edu.au}
\affiliation{%
  \institution{The University of Adelaide}
  \city{Adelaide}
  \country{Australia}
}
\author{Mingming Gong}
\email{mingming.gong@unimelb.edu.au}
\affiliation{%
  \institution{The University of Melbourne}
  \city{Melbourne}
  \country{Australia}
}
\affiliation{%
  \institution{MBZUAI}
  \city{Abu Dhabi}
  \country{United Arab Emirates}
}
\author{Lina Yao}
\email{lina.yao@data61.csiro.au}
\affiliation{%
  \institution{CSIRO’s Data 61}
  \institution{The University of New South Wales}
  \city{Sydney}
  \country{Australia}
}

\renewcommand{\shortauthors}{Xinshu Li, Ruoyu Wang, Erdun Gao, Mingming Gao, \& Lina Yao}

\begin{abstract}

Prompt learning has garnered attention for its efficiency over traditional model training and fine-tuning. However, existing methods, constrained by inadequate theoretical foundations, encounter difficulties in achieving causally invariant prompts, ultimately falling short of capturing robust features that generalize effectively across categories. To address these challenges, we introduce the $\textit{\textbf{DiCap}}$ model, a theoretically grounded $\textbf{Di}$ffusion-based $\textbf{C}$ounterf$\textbf{a}$ctual $\textbf{p}$rompt learning framework, which leverages a diffusion process to iteratively sample gradients from the marginal and conditional distributions of the causal model, guiding the generation of counterfactuals that satisfy the minimal sufficiency criterion. Grounded in rigorous theoretical derivations, this approach guarantees the identifiability of counterfactual outcomes while imposing strict bounds on estimation errors. We further employ a contrastive learning framework that leverages the generated counterfactuals, thereby enabling the refined extraction of prompts that are precisely aligned with the causal features of the data. Extensive experimental results demonstrate that our method performs excellently across tasks such as image classification, image-text retrieval, and visual question answering, with particularly strong advantages in unseen categories.
\end{abstract}

\ccsdesc[500]{Computing methodologies~Computer vision tasks}

\keywords{Vision Language Models, Prompt Learning, Diffusion Process, Counterfactual Generation}

\maketitle

\section{Introduction}
Pre-trained vision-language foundation models \citep{radford2021learning,li2022blip,li2023blip} integrate large-scale multimodal information, effectively breaking down boundaries between different domains and significantly broadening their applications. To facilitate knowledge transfer from these models to downstream tasks, prompt engineering \citep{wei2022chain,shin2020autoprompt} employs optimized templates that guide the model in generating more relevant and accurate outputs. However, the manual creation of prompts is time-consuming and labor-intensive, leading to the emergence of prompt learning \citep{khattak2023maple,zhou2022conditional,zhou2022learning}, which allows for the automatic acquisition of tunable prompts and improves the efficiency of transferring knowledge to downstream tasks.

\begin{figure}[t]
  \centering
   \includegraphics[width=0.8\linewidth]{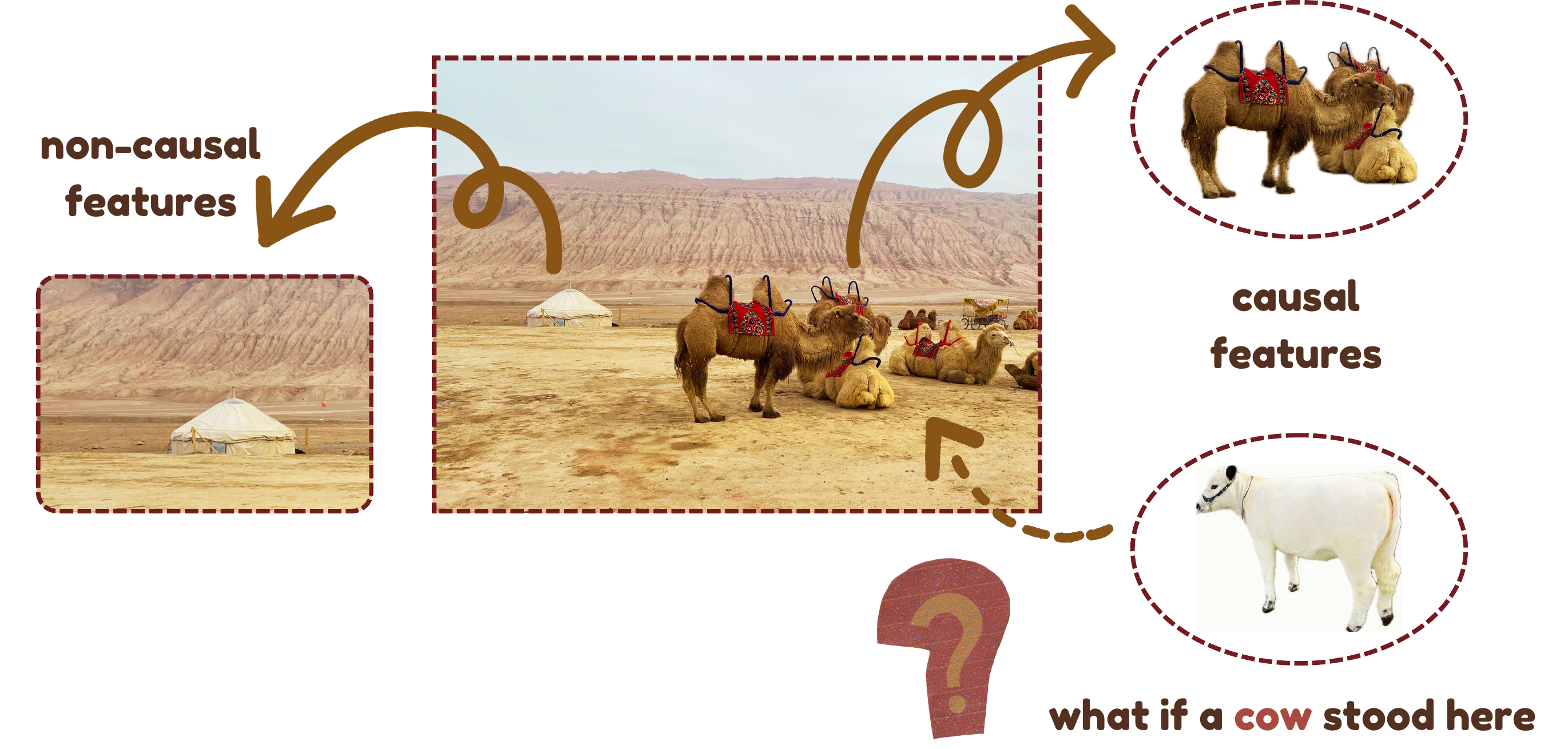}
   \caption{\textbf{A motivating example:} In camel classification tasks, robust prompt embeddings should align with the camel's causal features, such as physical traits, rather than non-causal features, like desert backgrounds or yurts. Counterfactual learning poses questions like “What if a cow stood here?” to generate counterfactual images, which are used in contrastive learning framework to guide prompts in identifying and aligning with causal features, offering a practical approach to causal-invariant prompt embeddings.}
   \label{fig:camel}
\end{figure}

Current prompt learning methods often suffer from capturing spurious correlations between variables, especially in the absence of additional constraints \citep{liu2023pre,liu2023kept}, which can lead to degraded performance under distribution shifts. To mitigate this issue, causal prompt learning methods \citep{fu2020counterfactual,he2022cpl,li2022supporting,lyu2023psychologically} have been proposed to identify and leverage underlying causal relationships, thereby improving the robustness and generalization of prompts. As shown in Figure \ref{fig:camel}, prompt embeddings should focus on the causal features of an image, such as the camel's hump, rather than the non-causal factors like the frequently co-occurring desert background or yurts. However, an important challenge remains: how can prompt learning effectively distinguish causal features from spurious correlations in complex visual data?

 Counterfactual learning \citep{swaminathan2015self,goyal2019counterfactual,johansson2016learning, li2022contrastive,li2024distribution,li2024self} offers a promising approach to this challenge, which involves answering the question of how the outcomes of an event would change if its underlying causal factors were modified \citep{pearl2009causality}. Specifically, in the case of Figure \ref{fig:camel}, the counterfactual speculation might be: ``What would the image look like if a cow were standing where the camel originally stood?" By generating counterfactual images as negative samples and aligning the original image as a positive sample, contrastive learning encourages prompt embeddings to capture consistent features across variations in non-causal factors, guiding them to focus on essential causal features for robust visual understanding.

Nonetheless, generating counterfactual samples remains a significant challenge, particularly for high-dimensional image data \citep{pawlowski2020deep,yang2021causalvae}. Current efforts primarily rely on multimodal information \citep{he2022cpl,li2022supporting,li2025contrastive}, such as identifying semantic similarities between related image prompts or leveraging knowledge graphs to construct counterfactuals. However, these approaches face critical limitations: 1) Their success hinges on sufficient multimodal information, which is often scarce and poses a significant bottleneck. 2) The lack of a robust theoretical foundation to ensure counterfactual identifiability and constrain estimation errors undermines their reliability. 3) The use of low-dimensional vectors to represent counterfactual images limits the model's ability to capture shared high-dimensional non-causal features and distinguish causal ones. Therefore, a key challenge lies in developing methods that generate high-dimensional counterfactual samples with guaranteed error bounds while reducing dependence on multimodal information to enhance the robustness of prompt learning.

In this work, we leverage recent advancements in diffusion models \citep{songdenoising,ho2020denoising,dhariwal2021diffusion,sanchez2022diffusion} to design methods for counterfactual prompt learning. Diffusion models offer several key advantages for this task: 1) Their isometric diffusion process preserves high-dimensional image features, minimizing information loss and enabling the possibility of uni-modal reliance; 2) The iterative sampling mechanism \citep{ho2020denoising} naturally extends to interventions on key causal features \citep{chao2023interventional}, facilitating the minimal sufficiency of the generated counterfactual images; 3) Diffusion models are underpinned by rigorous mathematical guarantees, ensuring the reliability of counterfactual generation.

Specifically, we propose a novel \textbf{Di}ffusion-based \textbf{C}ounterf\textbf{a}ctual \textbf{p}rompt (\textbf{\textit{DiCap}}) learning framework to align prompts with causal invariant features. First, we invert the diffusion process via denoising \citep{ho2020denoising}, approximating the gradient of the log-likelihood of the input distribution. In parallel, we leverage anti-causal predictors \citep{scholkopf2012causal}, which infer causes from observed effects by reversing the natural cause-effect flow. Subsequently, causal interventions are performed using gradients derived from these predictors, enabling minimal sufficiency of the generated counterfactuals via Markov Chain Monte Carlo sampling. Furthermore, we provide rigorous theoretical derivations to establish the error bounds of counterfactual estimation, ensuring its reliability. Finally, we design a contrastive learning task \citep{chen2020simple,khosla2020supervised} to train tunable prompt embeddings, enabling them to attract factual samples while repelling counterfactual ones, effectively aligning with the causal features in the images.

Our main contributions are summarized as follows:

\begin{itemize}
\item We propose a novel, theoretically grounded \textbf{Di}ffusion-based \textbf{C}ounterf\textbf{a}ctual \textbf{p}rompt (\textbf{\textit{DiCap}}) learning framework that generates robust prompt representations aligned with causal features in images, significantly improving model performance on unseen distributions.  
 
\item Through rigorous theoretical derivations, we established the sufficient conditions that guarantee the identifiability of the generated counterfactual samples and bound the counterfactual estimation errors.
\item Extensive experimental results on image classification, image-text retrieval, and visual question answering (VQA) consistently confirm the superiority, stability, and generalizability of our approach over existing baselines.
\end{itemize}

\section{Related Work}
\label{sec:related}

\subsection{Causal Prompt Learning} 

Prompt learning has evolved from static prompts in CLIP \citep{radford2021learning} to learnable soft prompts in CoOp \citep{zhou2022learning}, improving task-specific adaptation but struggling with unseen classes due to distribution shifts. CoCoOp \citep{zhou2022conditional} mitigates this by conditioning prompts on image features, enhancing generalization further. Recent advancements highlight the potential of causal learning in refining prompt robustness. \citet{li2024ccprefix} leverages causal contrastive learning to generate instance-dependent soft prefixes for multi-class classification, addressing label ambiguity. \citet{zhang2024causal} introduces causal prompting with front-door adjustment to reduce biases in large language models. \citet{li2022supporting} presents CPKP, which improves semantic richness and generalization through ontological knowledge graphs. \citet{lyu2023psychologically} explores causal relations in sentiment classification using three causal prompts. \citet{li2025contrastive} combines counterfactual generation and learnable prompts for better radiology report generation. \citet{he2022cpl} proposes a method that constructs counterfactuals by identifying semantically similar images. 

Despite these advancements, existing methods lack theoretical guarantees for counterfactual generation and fail to fully integrate causal reasoning into prompt learning. To bridge this gap, we propose DiCap, a causal diffusion model for prompt learning that establishes theoretical error bounds in counterfactual generation, improving both effectiveness and generalization to unseen classes.

\begin{figure*}[t]
  \centering
   \includegraphics[width=\linewidth]{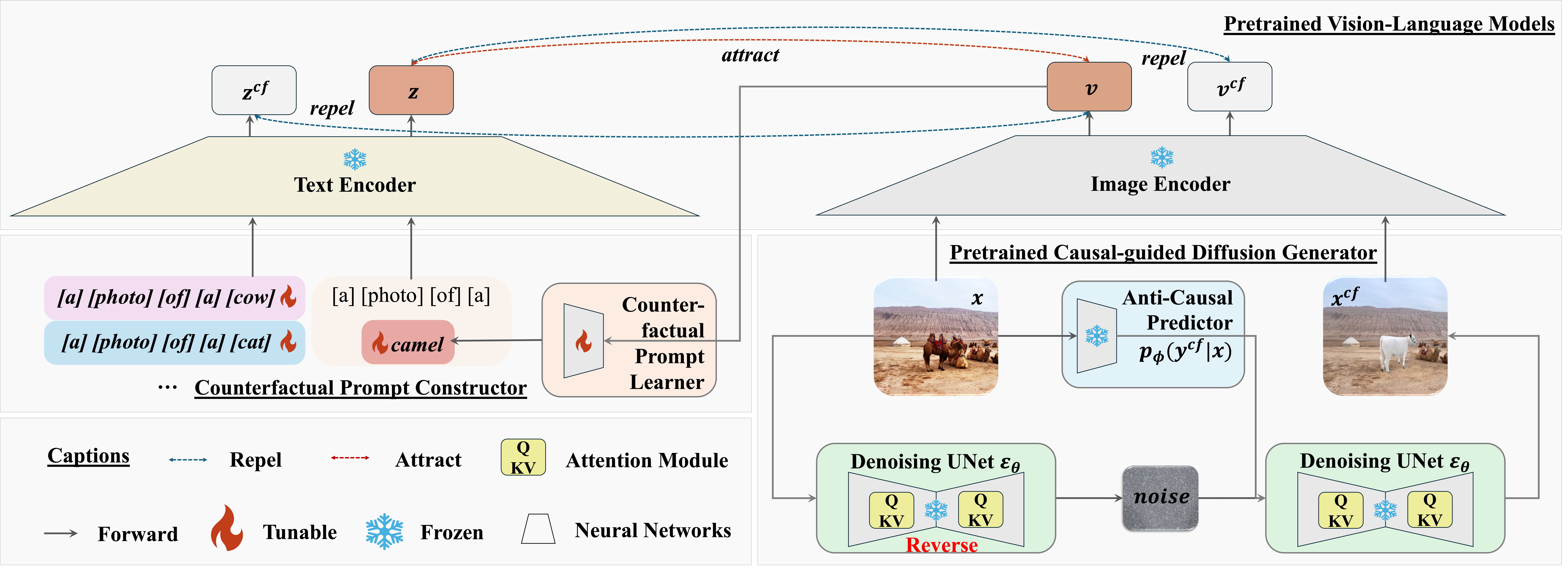}

   \caption{\textbf{DiCap Method Overview:} The process begins by inputting the factual image \(x\) into a pretrained denoising model, generating noise as a proxy for exogenous variables. The image is also passed through an anti-causal predictor to obtain the counterfactual label gradient  \( {\nabla}_{\boldsymbol{x}} p_\phi(y^{cf} | \boldsymbol{x}) \), which guides the denoising model to generate the counterfactual image \(\boldsymbol{x}^{cf}\). Each factual image is paired with a unique trainable dynamic prompt vector that captures causal features using a dual-contrastive loss. In addition to the standard contrastive loss, a “counterfactual as hardest” strategy aligns the dynamic prompt embedding with the factual image while distancing it from the counterfactual. Fine-tuning is applied only to the trainable counterfactual prompt learner, with the remaining network components frozen for computational efficiency.}
   \label{fig:dicap}
\end{figure*}

\begin{figure}[t]
  \centering

   \includegraphics[width=0.7\linewidth]{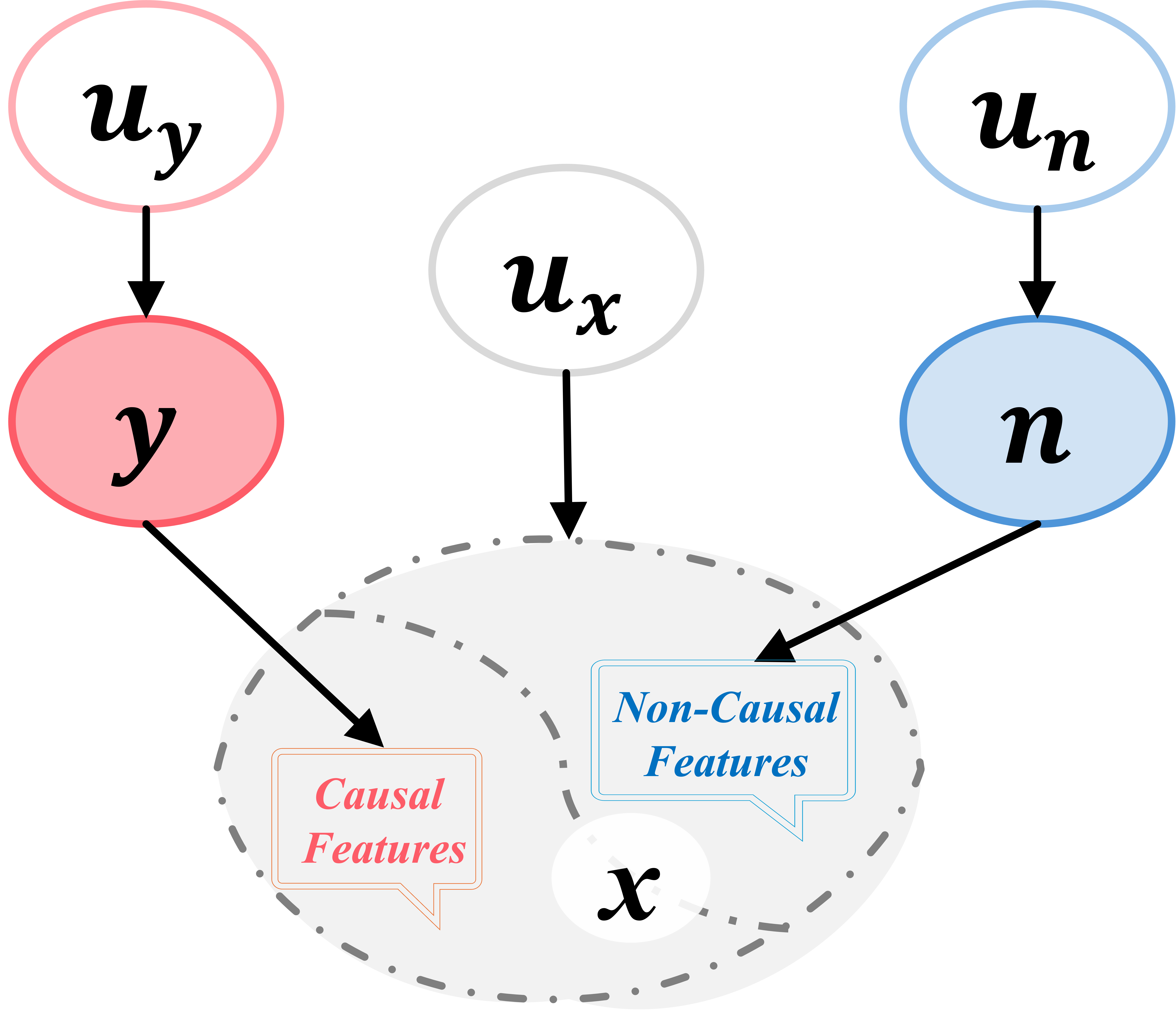}

   \caption{\textbf{The Data Generation Process} of the images: The features of the image \(\boldsymbol{x}\) can be divided into causal and non-causal components. The label of the image, \(\boldsymbol{y}\), serves as the cause of the causal features, while the remaining factors, \(\boldsymbol{n}\), cause the non-causal features. Additionally, \(\boldsymbol{u}_x\), \(\boldsymbol{u}_y\), and \(\boldsymbol{u}_n\) represent the exogenous variables which are the causes of these endogenous variables.}
   \label{fig:dgp}
\end{figure}
\subsection{Counterfactual Generation}
By constructing hypothetical scenarios with minimal causal modifications, counterfactual generation serves as a powerful tool for understanding causal mechanisms and improving model robustness. Existing counterfactual generation approaches fall into four categories: 1) Multimodal-guided methods: \citet{prabhu2023lance,kim2023grounding} leverage language models or textual concepts to reduce annotation costs but risk semantic misalignment; 2) Object-centric methods: \citet{zemni2023octet,jacob2022steex,jeanneret2023adversarial} enable fine-grained edits via scene priors but require external annotations and exhibit limited generalization; 3) Adversarial attack derivatives: \citet{khorram2022cycle} convert perturbations into semantic edits but prioritize robustness over causal interpretation. 4) Diffusion-based methods: Diffusion models' iterative sampling process aligns more naturally with causal interventions compared to other generative models. Early works \citep{weng2025fast,augustin2022diffusion,augustin2024dig,jeanneret2022diffusion} focused on generating high-fidelity samples using gradient guidance or regularization but were primarily designed for image editing, lacking theoretical guarantees for causal consistency. To incorporate counterfactual reasoning, \citet{sanchez2022diffusion} introduced classifier-guided diffusion models \citep{dhariwal2021diffusion}, leveraging external classifiers to direct sample generation toward specific counterfactual outcomes. Alternatively, \citet{chao2023interventional} explored classifier-free models \citep{ho2022classifier}, adopting a different approach to counterfactual sample generation without explicit classifier constraints. While these methods advance generation quality and scenario adaptability, they fail to address causal identifiability and cross-task generalization. 

Our DiCap framework bridges these gaps by integrating counterfactual generation with prompt learning. Grounded in structural causal models, DiCap enables minimal sufficiency of the generated counterfactuals via a denoising process and extracts causally invariant prompts through contrastive learning. By enforcing error-bounded theoretical rigor and enabling multimodal task compatibility, DiCap surpasses existing methods reliant on heuristic designs or domain-specific priors.

\begin{figure*}[t]
  \centering

   \includegraphics[width=\linewidth]{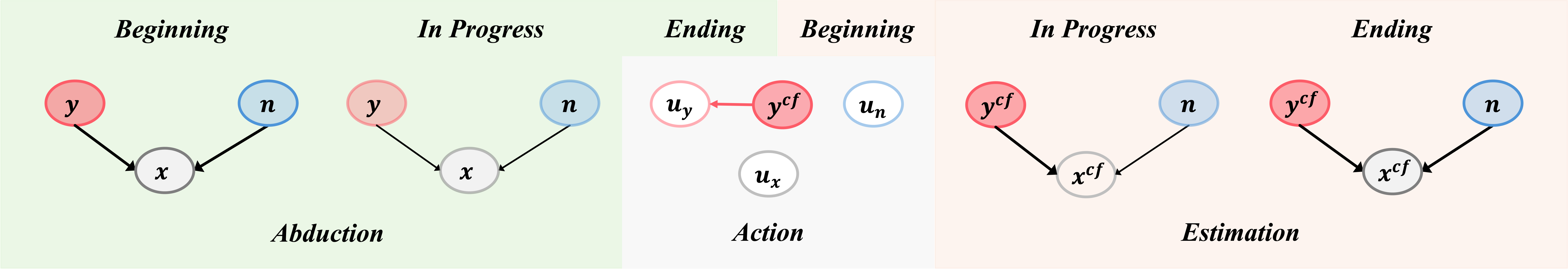}
   \caption{\textbf{Diffusion-Based Counterfactual Generation}: During diffusion, noise is progressively added to the factual image \(\boldsymbol{x}\), transforming endogenous variables governed by causal dependencies into independent exogenous variables (\textit{Abduction}). As noise increases, causal relationships weaken, represented by thinner arrows. By replacing the exogenous variable \(u_{y}\) which is the cause of the factual label \(y\) with the counterfactual label \(y^{cf}\) at each timestep (\textit{Action}), the denoising process, guided by the anti-causal gradient \( {\nabla}_{x} p_\phi(y^{cf} | x) \), generates counterfactual images \(\boldsymbol{x}^{cf}\) that minimize causal dependencies while enhancing non-causal features \(n\) (\textit{Estimation}).}
   \label{fig:causal_diff}
\end{figure*}
\section{Methodology}
\label{method}

\subsection{Method Overview}
As shown in Figure \ref{fig:dicap}, our DiCap model consists of the following key components: First, denoising sampling is performed using anti-causal gradient guidance from a pretrained diffusion model to generate counterfactual images, which are then used as the hardest negative samples in a dual-contrastive task to guide the generation of prompt embeddings that align with causal features. The following sections detail our framework and its theoretical foundations.

\subsection{Data Generation Process}
Figure \ref{fig:dgp} illustrates the general data generation process \citep{zimmermann2021contrastive} for the images. The features of the image $\boldsymbol x$ can be classified into causal and non-causal features. The label of the image, $y$, causes the causal features, while $\boldsymbol n$ represents other factors leading to non-causal features. Exogenous variables $u_x$, $u_y$, and $u_n$ represent the unobserved causes of $\boldsymbol x$, $y$, and $\boldsymbol n$, respectively.
\subsection{Diffusion-Based Counterfactual Generation}
\label{diff_model}

Following \citet{sanchez2022diffusion}, we generate counterfactuals via a classifier-guided diffusion model \citep{dhariwal2021diffusion}. However, while \citet{sanchez2022diffusion} interprets the model through a bivariate framework of images and labels, we adopt a more comprehensive perspective by incorporating non-causal factors into the reasoning process, refining the overall modeling of counterfactual variations.

Counterfactual reasoning involves abduction, action, and estimation. As shown in Figure \ref{fig:causal_diff}, we show how the diffusion model aligns with these stages to effectively generate counterfactuals, highlighting its unique advantages in counterfactual inference.

\textbf{Abduction:} The model identifies exogenous variables associated with nodes in the causal graph, such as \(\boldsymbol{u}_x\), \(\boldsymbol{u}_y\), and \(\boldsymbol{u}_n\), as shown in Figure \ref{fig:dgp}. This parallels the forward mechanism in diffusion models, where causal dependencies weaken, leading to independent exogenous variables \citep{sanchez2022diffusion}. We use a forward implicit diffusion process \citep{songdenoising}, reformulating the DDIM process as an ordinary differential equation (ODE) and applying the Euler approximation to derive independent exogenous factors.
\begin{equation}
\begin{split}
    \boldsymbol{x}_{t+1} \leftarrow &\sqrt{\bar{\alpha}_{t+1}}\left(\frac{\boldsymbol{x}_{t}-\sqrt{1-\bar{\alpha}_t} \epsilon_\theta\left(\boldsymbol{x}_{t}, t\right)}{\sqrt{\bar{\alpha}_t}}\right)\\
    &+\sqrt{1-\bar{\alpha}_{t+1}} \boldsymbol{\epsilon}_\theta\left(\boldsymbol{x}_{t}, t\right), t = 0, \dots, T-1, \\
\end{split}
\label{diff_reverse}
\end{equation}
 where $\boldsymbol{x}_{0}={\boldsymbol{x}}$,  \(\boldsymbol{\epsilon}_\theta\) is a pretrained model to estimate the noise introduced at each time-step $t$, $\bar{\alpha}
_t := \prod_{j=0}^t (1 - \beta_j)$ means a time-dependent variance. 

In contrast to the forward process during the pretraining of the diffusion, where independent Gaussian noise is directly added to the image, we input the $\boldsymbol{x}_{t} $ into ${\boldsymbol \epsilon}_{\theta}$ to generate noise that aligns with most of the original features in each timestep, which is employed in both the forward noise addition and reverse denoising processes for counterfactual generation. This structured noise reduces interference from random noise on non-target attributes, ensuring the counterfactuals closely align with the factuals in non-causal features. However, this noise inevitably retains certain causal features related to the label, which will be addressed and adjusted in the \textbf{Action} phase.

\textbf{Action and Estimation:} In the Langevin dynamics framework, the diffusion process is controlled by gradient guidance, which is particularly effective for counterfactual generation \citep{songdenoising}. Following \citet{sanchez2022diffusion}, we guide the sampling using the gradient of the anti-causal predictor \( {\nabla}_{x} p_\phi(y^{cf} | x) \), where \( y^{cf} \) is the counterfactual label. This gradient directs the process at each timestamp, ensuring counterfactual images \( \boldsymbol{x}^{cf} \) reflect the causal effects of label changes as shown in the sampling equation below.
\begin{equation}
\begin{split}
\begin{aligned}
& \epsilon \leftarrow \boldsymbol{\epsilon}_\theta\left(\boldsymbol{x}_t, t\right)-s \sqrt{1-\bar{\alpha}_t} \nabla_{\boldsymbol{x}_t} \log p_\phi\left(y^{cf} \mid \boldsymbol{x}_t\right) \\
& \boldsymbol{x}_{t-1} \leftarrow \sqrt{\bar{\alpha}_{t-1}}\left(\frac{\boldsymbol{x}_t-\sqrt{1-\bar{\alpha}_t} \epsilon}{\sqrt{\bar{\alpha}_t}}\right)+\sqrt{1-\bar{\alpha}_{t-1}} \epsilon, \\
&t = T, \dots, 1,\\
\end{aligned}
\end{split}
\label{eq:intervention}
\end{equation}
where \(s\) is a hyperparameter \citep{ho2022classifier} controlling the scale of anti-causal gradients. In Section \ref{ex:cf_ev}, \(s\) is analyzed to reveal the impact of counterfactual image quality on prompt learning performance.

Diffusion-Based Counterfactual Generation offers the significant advantage of enabling the minimal sufficiency of the generated counterfactual images, which are defined as follows:

\begin{definition}
\textbf{Minimal sufficiency} of counterfactual images mandates that the counterfactual \(\boldsymbol{x}^{cf}\) is produced by the smallest perturbation to the factual image \(\boldsymbol{x}\) that alters its label from \(y\) to \(y^{cf}\). Formally,  
        \begin{equation}
       \boldsymbol{x}^{cf} = \arg\min_{\boldsymbol{x'}}\, d(\boldsymbol{x}, \boldsymbol{x'}) \ \text{subject to} \ k(\boldsymbol{x'}) = y^{cf} \ \text{and} \ y^{cf} \neq y,
        \label{minimal}
        \end{equation}
        where $d(\cdot, \cdot)$ is a distance metric and $k$ is a mapping function: ${\mathcal{X}} \to \mathcal{Y}$.
    \label{def:minimal_suffi}
\end{definition}
\textbf{\textit{Remark:}} Reflecting on our counterfactual generation mechanism in Figure \ref{fig:causal_diff}, we identify three key aspects that ensure the minimal sufficiency of counterfactual images. 1) Forward noise injection weakens causal dependencies, generating independent noise and plausible latent exogenous variables \(\boldsymbol{u_x}\), \(\boldsymbol{u_n}\), and \(\boldsymbol{u_y}\). 2) The anti-causal predictor gradient \( {\nabla}_{x} p_\phi(y^{cf} | x) \) guides the generation process, aligning the image with the counterfactual class $y^{cf}$. 3) Noise derived from the pretrained diffusion model \(\epsilon_\theta\) governs both forward perturbation and reverse denoising, preserving the relationship between non-causal features and the counterfactual image \(\boldsymbol{x}^{cf}\). Together, these mechanisms ensure \textbf{minimal yet sufficient} changes during the transition.

\subsection{Bounding Counterfactual Error}

We begin by articulating the necessary conditions for recovering the latent exogenous variable via an invertible function, as shown in Theorem \ref{theory_1}. These conditions are derived through rigorous mathematical reasoning within the structural causal models (SCM) \citep{pearl2009causality} discussed in this paper. 

\begin{theorem}

Suppose that for \( \boldsymbol{x} \in \mathcal{X} \subset \mathbb{R}^m \) and continuous exogenous noise \( \boldsymbol{u}_x \sim \textup{Unif}[0, 1]^m \) with \( m \geq 3 \), \( \boldsymbol{x} \) is governed by the structural equation:
\begin{equation}
    \boldsymbol{x} = f(y, \boldsymbol{n}, \boldsymbol{u}_x),
    \label{eq:theo_1}
\end{equation}
where \( y \in \mathcal{Y} \subset \mathbb{R} \), \( \boldsymbol{n} \in \mathcal{N} \subset \mathbb{R}^m \), and \( \boldsymbol{u}_x \perp y, \boldsymbol{n} \). Consider a forward function \( g : \mathcal{X} \to \mathcal{L} \) and a reverse function \( h : \mathcal{L} \times \mathcal{N} \times \mathcal{Y} \to \mathcal{X} \), such that:
\begin{equation}
    l := g(\boldsymbol{x}), \quad \hat{\boldsymbol{x}} := h(l, y, \boldsymbol{n}).
\end{equation}
Assume the following conditions hold:
\begin{enumerate}
    \item The structural function \( f \) is invertible and differentiable with respect to \( \boldsymbol{u}_x \), and its Jacobian \( J_{f_{y, \boldsymbol{n}}} \) is positive definite for all \( y \in \mathcal{Y} \) and \( \boldsymbol{n} \in \mathcal{N} \).
    \item The function \( g \) is invertible and differentiable with respect to \( \boldsymbol{x} \).
    \item The recovered latent variable is independent of the parent variables, \( g(\boldsymbol{x}) \perp y, \boldsymbol{n} \).

    \item The transformation \( q_{y, \boldsymbol{n}}(\boldsymbol{u}_x) := g(f(y, \boldsymbol{n}, \boldsymbol{u}_x)) \) satisfies \( J_{q_{y, \mathbf{n}} \mid q_{y, \mathbf{n}}^{-1}(\mathbf{l})} = c(y, \mathbf{n}) A, \quad \text{for all } \mathbf{l} \in \mathcal{L} \text{ and } y \in \mathcal{Y}, \, \mathbf{n} \in \mathcal{N}
 \), where \( c \) is a scalar function and \( A \) is an orthogonal matrix.
\end{enumerate}
Then, \( g(f(y, \boldsymbol{n}, \boldsymbol{u}_x)) = \tilde{q}(\boldsymbol{u}_x) \) for some invertible function \( \tilde{q} \).
\label{theory_1}
\end{theorem}
The first condition in \textbf{Theo.} \ref{theory_1} ensures \textbf{\textit{identifiability}} \citep{lu2020sample,nasr2023counterfactual} of causal estimation under additive noise models \citep{hartford2017deep} and post-nonlinear models \citep{zhang2012identifiability}, from which corollaries are derived to identify settings for precise counterfactual estimation and establish error bounds.

\begin{corollary}
Suppose the function pairs $(g,h)$ that satisfy $h(g(\boldsymbol{x}),y,\boldsymbol{n})=\boldsymbol{x}$ where $\boldsymbol{x} := f(y,\boldsymbol{n},\boldsymbol{{u}_{x}})$ and a do-operation $do(\boldsymbol{y}:=y^{cf})$, then counterfactual estimation $h(g(\boldsymbol{x}),y^{cf},\boldsymbol{n})$ will recover true counterfactual outcome $\boldsymbol{x^{cf}}:=f(y^{cf},\boldsymbol{n},\boldsymbol{u_x})$ under the conditions in \textbf{Theo.} \ref{theory_1}.
    \label{method:corol_1}
\end{corollary}

\textbf{Corol.} \ref{method:corol_1} implies that the counterfactual predictions will align with the true counterfactual outcomes if no information is lost in the forward and reverse steps. Building on this, we further derive the following error bound for classifier-guided diffusion-based counterfactual generation.

\begin{corollary}
Consider the function pairs $(g,h)$, if the reconstruction error is less than $\delta$, i.e., $d(h(g(\boldsymbol{x}),y,\boldsymbol{n}),\boldsymbol{x})\leq \delta$, then the error between counterfactual estimation $h(g(\boldsymbol{x}),y^{cf},\boldsymbol{n})$ and true counterfactual outcome $\boldsymbol{x^{cf}}:=f(y^{cf},\boldsymbol{n},\boldsymbol{u_x})$ will be at most $\delta$, i.e., $d(h(g(\boldsymbol{x}),y^{cf},\boldsymbol{n}),\boldsymbol{x^{cf}})\leq \delta$, under the conditions in \textbf{Theo.} \ref{theory_1}.
    \label{method:corol_2}
\end{corollary}

\textbf{Corol.} \ref{method:corol_2} indicates that if the similarity between the reconstructed and factual images is high, the model can yield credible counterfactual samples with substantial precision. \textbf{Theo.} \ref{theory_1}, \textbf{Corol.} \ref{method:corol_1} and \textbf{Corol.} \ref{method:corol_2} provide the theoretical guarantees for the counterfactual image generation method based on classifier-guided diffusion models.

\subsection{Counterfactual Prompt Construction}
\label{causal_prompt_construct}

\textbf{Closest as Counterfactual:} We then discuss how to select counterfactual labels to generate images that enhance prompt learning by aligning prompt embeddings with the causal features in the images rather than redundant details. Following \citet{robinson2020contrastive}, harder-to-distinguish negative samples are more effective as they encourage contrastive learning to capture subtle class distinctions. Therefore, we select the class with the second closest predicted probability to the factual label \( y \) as the counterfactual label \( y^{cf} \). For instance, for an image labeled as a cat, when input into the anti-causal predictor, the probability of predicting it as a cat should be closest to that of another feline, such as ``tiger", rather than an unrelated animal like ``dog". Thus, the gradients related to ``tiger" is used to synthesize the counterfactual image, which subsequently functions as the most challenging negative sample in the contrastive learning task. Experimental results in Section \ref{ex:sample} demonstrate the effectiveness of this sampling strategy.

\textbf{Counterfactual as Hardest:} Our prompt learning method is built upon dual contrastive learning tasks. The first adopts the vision-language contrastive paradigm from \citet{radford2021learning}, implementing the Conditional Context Optimization loss (Eq. \eqref{cocoop}) where image features anchor a similarity space that attracts their paired text prompts while repelling counterfactual class prompts. 
\begin{equation}
\mathcal{L}_{basic}=\sum\limits^{m}\limits_{i=1} - \log{\dfrac{exp(\boldsymbol{v}^{i}\cdot{ g({\boldsymbol w}^{y^i}(\boldsymbol{v}^{i}))/{\tau})}}{\sum\limits^{m}\limits_{j=1}exp(\boldsymbol{v}^{i}\cdot{g({\boldsymbol w}^{y^j}(\boldsymbol{v}^{i}))}/{\tau})}},
\label{cocoop}
\end{equation}
where $\boldsymbol v$ denotes the image embeddings learned by the image encoder of pretrained vision language model (e.g., CLIP). $i$, $m$ and $\tau$ indicate $i_{th}$ image in the training dataset, training data size and temperature parameter. $ g(\cdot)$ denotes the text encoder and $\boldsymbol w$ represents the tunable prompt embeddings, which are learned by the Counterfactual Prompt Constructor.

The second contrastive task uses generated counterfactuals: prompt embeddings serve as anchors, factual image embeddings as positive samples, and counterfactual embeddings as negative samples. The objective function is:
\begin{equation}
\begin{split}
    \mathcal{L}_{cf}=& \sum\limits^{m}\limits_{i=1} - \log exp(\boldsymbol{v}^{i}\cdot{\boldsymbol g({\boldsymbol w}^{y^i}(\boldsymbol{v}^{i}))/{\tau})}\\
    &+ \log (exp(\boldsymbol{v}^{i}\cdot{\boldsymbol g({\boldsymbol w}^{y^i}(\boldsymbol{v}^{i}))}/{\tau})+exp((\boldsymbol{v}^{cf})^{i}\cdot{\boldsymbol g({\boldsymbol w}^{y^i}(\boldsymbol{v}^{i}))}/{\tau})),\\
\end{split}
\label{contra_diff}
\end{equation}
where $(\boldsymbol{v}^{cf})^{i}$ refers to the embedding of counterfactual image related to factual image $\boldsymbol{x}^{i}$.

In summary, the overall objective function for the \textbf{\textit{Di}}ffusion-based \textbf{C}ounterf\textbf{a}ctual \textbf{\textit{p}}rompt learning (\textbf{\textit{DiCap}}) method is formulated as follows.
\begin{equation}
\begin{split}
    \mathcal{L}_{total}= \mathcal{L}_{basic} +  \lambda \mathcal{L}_{cf},
\end{split}
\label{eq:total_loss}
\end{equation}
where $\lambda$ is an adjustable hyperparameter. Pseudo-code of \textbf{Dicap} is provided in the Algorithm \ref{algo:dicap}.

\begin{algorithm}[htbp]
\caption{Diffusion-based Counterfactual Prompt Learning   }
\begin{algorithmic}
   \STATE {\bfseries Input:} Training data size $m$, Training dataset $\mathcal{D}=\{(\boldsymbol{v}^i,y^i)|i \in {1,2,...,m}\}$, Counterfactual dataset $\mathcal{D}^{cf}=\{(\boldsymbol{v}^i,({\boldsymbol{v}^{cf}})^i)|i \in {1,2,...,m}\}$, CLIP pretrained text encoder $\boldsymbol{g}(\cdot)$, tunable prompt embedding $\boldsymbol{w}$, temperature $\tau$, loss-re-weight hyper-parameter $\lambda$.
   \STATE {\bfseries Output:} Prompt embedding $\boldsymbol{w}$ after trained.
   \STATE Initialize $\boldsymbol{w}$
   \FOR{$i=0$ {\bfseries to} $m-1$}
   \STATE  {\color{gray}{\# Calculate basic loss:}}
   \STATE   $\mathcal{L}_{basic} += - \log{\dfrac{exp(\boldsymbol{v}^{i}\cdot{\boldsymbol g({\boldsymbol w}^{y^i}(\boldsymbol{v}^{i}))/{\tau})}}{\sum\limits^{m}\limits_{j=1}exp(\boldsymbol{v}^{i}\cdot{\boldsymbol g({\boldsymbol w}^{y^j}(\boldsymbol{v}^{i}))}/{\tau})}}$
   \STATE  {\color{gray}{\# Calculate counterfactual loss:}}
   \STATE   $    \mathcal{L}_{cf} += \sum\limits^{m}\limits_{i=1} - \log exp(\boldsymbol{v}^{i}\cdot{\boldsymbol g({\boldsymbol w}^{y^i}(\boldsymbol{v}^{i}))/{\tau})}+ \log (exp(\boldsymbol{v}^{i}\cdot{\boldsymbol g({\boldsymbol w}^{y^i}(\boldsymbol{v}^{i}))}/{\tau})+exp((\boldsymbol{v}^{cf})^{i}\cdot{\boldsymbol g({\boldsymbol w}^{y^i}(\boldsymbol{v}^{i}))}/{\tau}))$
   \STATE  {\color{gray}{\# Calculate total loss:}}
   \STATE $    \mathcal{L}_{total}= \mathcal{L}_{basic} +  \lambda \mathcal{L}_{cf}$
   \STATE Update $\boldsymbol{w}$ according to $\mathcal{L}_{total}$
   \ENDFOR
   \STATE return $\boldsymbol{w}$
\end{algorithmic}
\label{algo:dicap}
\end{algorithm}

\begin{table*}[t]
    \centering
    \vskip 0.15in
    \renewcommand\arraystretch{1.3}
    \begin{tabular}{ccccccccc}
    \hline
       \textbf{\textit{ Seen}} & \textbf{Caltech101} & \textbf{OxfordPets} & \textbf{Flowers102} & \textbf{Food101} & \textbf{StanfordCars} & \textbf{Sun397} & \textbf{ImageNet} & \textbf{Average}  \\ \hline
         \textbf{CLIP }& 90.64	&91.12	&69.80	&83.10	&55.45	&66.72	&70.30	&75.30 \\ \hline
        \textbf{CoOp} & \textbf{97.93} \textcolor{green}{8.04} &92.82 \textcolor{green}{1.87}	&\textbf{96.15} \textcolor{green}{37.8}&90.27 \textcolor{green}{8.63}&\textbf{74.76} \textcolor{green}{34.8}&	81.24 \textcolor{green}{21.8}&	86.25 \textcolor{green}{22.7}	&88.49 \textcolor{green}{17.5} \\ 
     \textbf{CoCoOp} & 97.74 \textcolor{green}{7.83}&	95.43 \textcolor{green}{4.73}&94.78 \textcolor{green}{35.8}	&90.76	\textcolor{green}{9.22}&	71.14 \textcolor{green}{28.3}	&79.73	\textcolor{green}{19.5}	&86.13 \textcolor{green}{22.5}	&87.96 \textcolor{green}{16.8} \\ \hline
     \textbf{CPL} &97.55	\textcolor{green}{7.62}	&95.27	\textcolor{green}{4.55}	&94.11 \textcolor{green}{34.8}	&90.73 \textcolor{green}{9.18}	&70.54	\textcolor{green}{27.2}	&79.47	\textcolor{green}{19.1} &\textbf{86.35}	\textcolor{green}{22.8}	&87.72 \textcolor{green}{16.5} \\ \hline
     
       \textbf{ Dicap} & 97.87	\textcolor{green}{7.98}	&\textbf{95.66} \textcolor{green}{4.98}	&94.12	\textcolor{green}{34.8}	&\textbf{90.91}	\textcolor{green}{9.40}	&73.07	\textcolor{green}{31.8}	&\textbf{82.38}	\textcolor{green}{23.5}	&86.07	\textcolor{green}{22.4}	&\textbf{88.58}	\textcolor{green}{17.6} \\ 
     \hline
     
 \hline
       \textbf{\textit{ Unseen}}  &\textbf{Caltech101} & \textbf{OxfordPets} & \textbf{Flowers102} & \textbf{Food101} & \textbf{StanfordCars} & \textbf{Sun397} & \textbf{ImageNet} & \textbf{Average}  \\ \hline
        \textbf{CLIP}& 92.16&94.13&\textbf{74.26}&91.20&	73.65&70.52&77.97&81.98 \\ \hline
        \textbf{CoOp} & 89.74 \textcolor{red}{2.60}	&95.41 \textcolor{green}{1.36}	&72.00 \textcolor{red}{3.04}&91.00 \textcolor{red}{0.22}&73.68 \textcolor{green}{0.04}	&73.45 \textcolor{green}{4.16}	&79.92 \textcolor{green}{2.50}	&82.17 \textcolor{green}{0.23} \\ 
     \textbf{CoCoOp} & 90.90	\textcolor{red}{1.37}	&94.87 \textcolor{green}{0.79}&66.21 \textcolor{red}{10.8}&91.61 \textcolor{green}{0.45}	&72.74	\textcolor{red}{1.24}	&72.48 \textcolor{green}{2.78}	&82.98 \textcolor{green}{6.43}	&81.68 \textcolor{red}{0.37} \\ \hline
         \textbf{CPL} & 92.90 \textcolor{green}{0.80}&97.93 \textcolor{green}{4.04}&72.62	\textcolor{red}{2.21}	&91.62 \textcolor{green}{0.46}&73.04	\textcolor{red}{0.83}&76.77	\textcolor{green}{8.86}&84.02 \textcolor{green}{7.76}&84.13 \textcolor{green}{2.62}\\ \hline
       \textbf{ Dicap} & \textbf{95.09} \textcolor{green}{3.18}&\textbf{98.01} \textcolor{green}{4.12}	&73.61 \textcolor{red}{0.88}	&\textbf{91.84} \textcolor{green}{0.70}&\textbf{74.92} \textcolor{green}{1.72}&\textbf{77.88} \textcolor{green}{10.4}&\textbf{84.77} \textcolor{green}{7.95}&\textbf{85.07} \textcolor{green}{3.87} \\ 
     \hline

    \end{tabular}
\caption{\textbf{Performance comparison} of accuracy rate between $DiCap$ and the SOTA baselines on the \textbf{image classification task}. Bold indicates the method with the best performance. The \textcolor{green}{green} numbers represent the performance improvement over the CLIP model, while the \textcolor{red}{red} numbers indicate the performance degradation relative to CLIP. The same applies to the Table \ref{itr_vqa}.}
\label{classification_result}
\end{table*}

\begin{table*}[ht]
\centering
\begin{minipage}[b]{0.6\textwidth}
\centering
\renewcommand\arraystretch{1.3}
\begin{tabular}{ccccc}
\hline
\textbf{Training} &\multirow{2}{*}{\textbf{Methods}}&\multirow{2}{*}{\textbf{Flickr30k}}&\multirow{2}{*}{{\textbf{MSCOCO}}}&\multirow{2}{*}{{\textbf{Average}}}\\
\textbf{Data Used} &&&&\\
\hline
0	&CLIP		&78.90			&67.70	&73.30\\
\hline
\multirow{3}{*}{\textbf{0.5\%}}	&CoCoOp	&76.80	\textcolor{red}{2.66	}&73.50 \textcolor{green}{8.57}&75.15	\textcolor{green}{2.52}\\
	&CPL	&84.00 \textcolor{green}{6.46}	&74.30 \textcolor{green}{9.75}&79.15 \textcolor{green}{7.98}\\
	&DiCap	&\textbf{85.80} \textcolor{green}{8.75}	&\textbf{75.10} \textcolor{green}{10.9}&\textbf{80.45} \textcolor{green}{9.75}\\
\hline
\multirow{3}{*}{\textbf{1\%}}&CoCoOp	&84.60 \textcolor{green}{7.22	}&74.33 \textcolor{green}{9.79}&79.47	\textcolor{green}{8.41}\\
	&CPL	&86.60 \textcolor{green}{9.76}&75.42 \textcolor{green}{11.4}&81.01 \textcolor{green}{10.5}\\
	&DiCap	&\textbf{87.20} \textcolor{green}{10.5	}&\textbf{75.82}	\textcolor{green}{12.0}&\textbf{81.51} \textcolor{green}{11.2}\\
\hline
\multirow{3}{*}{\textbf{1.5\%}}&CoCoOp	&84.90 \textcolor{green}{7.60	}&75.25 \textcolor{green}{11.2}&80.08	\textcolor{green}{9.24}\\
	&CPL	&85.90 \textcolor{green}{8.87	}&73.85 \textcolor{green}{9.08}&79.88 \textcolor{green}{8.97}\\
	&DiCap	&\textbf{86.20 }\textcolor{green}{9.25}&\textbf{75.95} \textcolor{green}{12.2}&\textbf{81.08} \textcolor{green}{10.6}\\
\hline
\end{tabular}
\subcaption{Image-Text Retrieval}
\label{itr}
\end{minipage}%
\begin{minipage}[b]{0.4\textwidth}
\centering
\renewcommand\arraystretch{1.3}
\begin{tabular}{cccc}
\hline
\textbf{Training} &\multirow{2}{*}{\textbf{Methods}}&\multicolumn{2}{c}{\textbf{VQAv2}}\\
\textbf{Data Used}&&\textbf{Seen}&\textbf{Unseen}\\
\hline
0	&CLIP	&16.01		&17.60	\\ \hline
\multirow{3}{*}{\textbf{0.1\%}}	&CoCoOp	&\textbf{23.14}	\textcolor{green}{44.5}&23.10	\textcolor{green}{31.3}\\
	&CPL	&22.00	\textcolor{green}{37.4}	&20.97	\textcolor{green}{19.1}\\
	&DiCap	&22.97	\textcolor{green}{43.5}&\textbf{23.87}	\textcolor{green}{35.6}\\
\hline
\multirow{3}{*}{\textbf{0.2\%}}	&CoCoOp	&35.60	\textcolor{green}{122}	&22.20 \textcolor{green}{26.1}\\ 
	&CPL	&35.08	\textcolor{green}{119}&21.60	\textcolor{green}{22.7}\\
	&DiCap	&\textbf{36.39}	\textcolor{green}{127}	&\textbf{23.10}	\textcolor{green}{31.3}\\ \hline
    \multirow{3}{*}{\textbf{0.5\%}}	&CoCoOp	&43.46	\textcolor{green}{171}	&22.60 \textcolor{green}{28.4}\\
	&CPL	&43.46	\textcolor{green}{171}	&20.40	\textcolor{green}{15.9}\\
	&DiCap	&\textbf{44.50}	\textcolor{green}{178}	&\textbf{25.50	}\textcolor{green}{44.9}\\ \hline
\end{tabular}
\subcaption{Visual Question Answering }
\label{vqa}
\end{minipage}
\caption{\textbf{Performance comparison} of Recall@1 (a) and accuracy rate (b) between $DiCap$ and the SOTA baselines on the \textbf{image-text retrieval} (a) and \textbf{visual question answering} tasks (b). }
\label{itr_vqa}
\end{table*}

\section{Experiments}
\label{ex}
\subsection{Task and Datasets}
\textbf{Image Classification:} Seven publicly available image classification datasets are used: Caltech101 \citep{griffin2007caltech}, OxfordPets \citep{parkhi2012cats}, Flowers102 \citep{nilsback2008automated}, Food101 \citep{bossard2014food}, StanfordCars \citep{krause20133d}, SUN397 \citep{xiao2010sun} and ImageNet \citep{deng2009imagenet}. The datasets are split into seen and unseen classes for generalisation assessment, with training conducted on the seen classes. Following the CLIP few-shot protocol, 16 training examples per class are used, and the full test set is employed for evaluation. The performance of the task is measured using the accuracy score.

\textbf{Image-Text Retrieval:} We evaluate image-text retrieval on MSCOCO \citep{lin2014microsoft} and Flickr30K \citep{plummer2015flickr30k} using the Karpathy split \citep{karpathy2015deep}: MSCOCO has 113K/5K/5K train/val/test images, and Flickr30K has 29K/1K/1K. For few-shot evaluation, training subsets use 0.5\%, 1\%, and 1.5\% of the data, with performance assessed on the whole test set. Results are measured using Recall at 1 (R@1).

\textbf{Visual Question Answering:} VQAv2 \citep{goyal2017making}, an extension of the VQA dataset \citep{antol2015vqa}, includes questions categorized as Number, Yes/No, and Other. Following \citet{anderson2018bottom}, VQA is treated as a classification task, where the model selects the correct answer from predefined options. Questions are converted into masked templates using a pre-trained T5 model \citep{raffel2023exploringlimitstransferlearning}, forming prompts that link questions and answers. The model predicts whether the prompt-image pair matches. For few-shot evaluation, 0.1\%, 0.2\%, and 0.5\% of instances are used for training. The performance of the task is measured using the accuracy score.

\subsection{Baselines}

We compare \textbf{DiCap} with: 1) the Zero-shot CLIP model \citep{radford2021learning}, 2) classical prompt learning methods CoOp and CoCoOp \citep{zhou2022learning, zhou2022conditional}, and 3) the state-of-the-art causal prompt learning method CPL \citep{he2022cpl}. All methods use CLIP as the pretrained vision-language model. The results for learning-based models are reported as the average over five independent runs to avoid randomness.

\subsection{Results}

\subsubsection{Accuracy Comparison}

This section presents the main results. Performance is measured relative to the zero-shot CLIP, with \textcolor{green}{green} representing the percentage of performance exceeding the CLIP baseline and \textcolor{red}{red} indicating the percentage falling below it.

\textbf{Image Classification:} From the experimental results in Table \ref{classification_result}, we have the following observations: 1) All prompt learning methods outperform zero-shot learning on seen classes; however, methods without causal learning, such as CoOp and CoCoOp, exhibit performance degradation on unseen classes. This suggests that these methods may have learned spurious correlations, underscoring the necessity of learning prompt representations with causal invariance. 2) Compared to CPL, another causality-based prompt learning method, our algorithm outperforms it across almost all seen classes and all unseen classes, demonstrating more robust predictive performance. This further validates that our counterfactual generation method better empowers prompt learning to align causal features in the images. 3) Our method achieves near-optimal performance across all datasets, with an average improvement of 17.6\% over CLIP on seen classes and 3.87\% on unseen classes, strongly affirming the superiority of our approach.

\textbf{Image Text Retrieval:} Table \ref{itr} shows the performance of diverse methods on the image text retrieval task. Our method performs optimally on unseen classes across three different few-shot learning settings. The vanilla instance-conditional prompt learning method, CoCoOp, experiences performance degradation on unseen classes of Flickr30k compared to zero-shot CLIP, further echoing our concerns. On the other hand, our DiCap model outperforms CPL in this task, demonstrating its ability to better focus on causal features and mitigate spurious correlations.

\begin{figure*}[t!]
    \centering
    \begin{minipage}[b]{0.32\textwidth}
      \centering
   \includegraphics[scale=0.345]{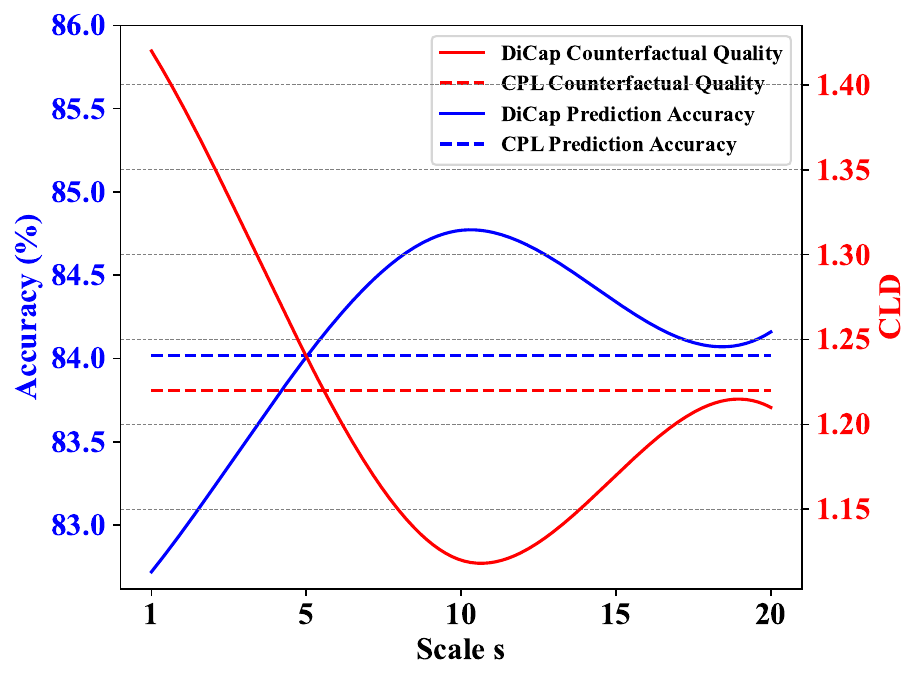}
   \subcaption{}
   \label{fig:scale}
    \end{minipage}
     \hfill
    \begin{minipage}[b]{0.32\textwidth}
        \centering
   \includegraphics[width=\linewidth]{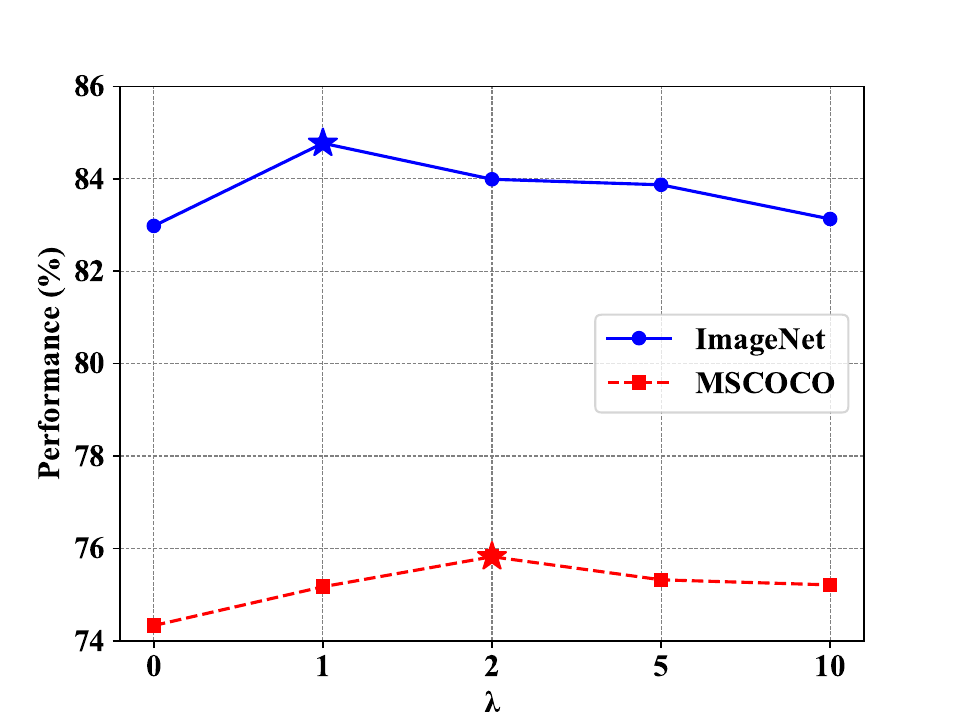}
   \subcaption{}
   \label{fig:lambda}
    \end{minipage}
    \hfill
    \begin{minipage}[b]{0.32\textwidth}
          \centering
   \includegraphics[width=\linewidth]{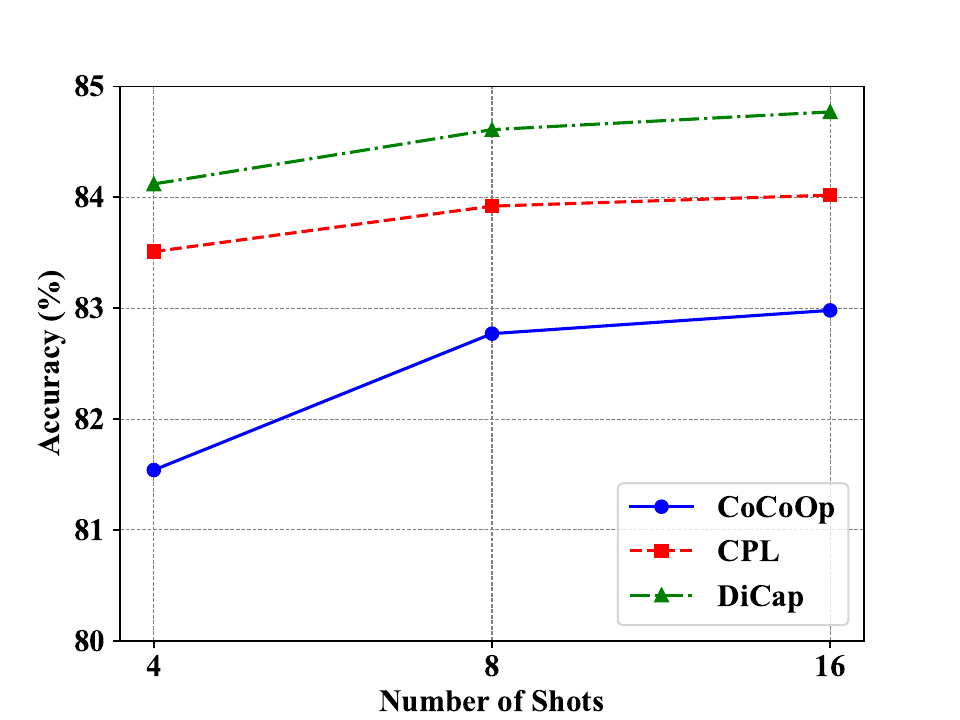}
   \subcaption{}
   \label{fig:shot}
    \end{minipage}
    \caption{(a) Counterfactuals enhance prompt learning. As the scale parameter \(s\) varies, model performance (left y-axis) and CLD scores (right y-axis) exhibit an inverse relationship. Since lower CLD scores indicate higher counterfactual quality, this highlights a strong positive correlation between counterfactual quality and model performance. Our DiCap method generates higher-quality counterfactuals within a broad range of \(5 < \text{scale} < 20\), which directly leads to improved predictive performance. (b) \textbf{Hyper-parameters analysis of} $\boldsymbol{\lambda}$ on ImageNet and MSCOCO datasets. The blue and red stars represent the best parameters for the datasets; (c) \textbf{three different shots} between $DiCap$ and two baselines on ImageNet unseen classes.} 
    \label{fig:side_by_side}
\end{figure*}

\textbf{Vision Question Answering:} In the VQA task, as shown in Table \ref{vqa}, the zero-shot CLIP model performs poorly, with an average accuracy rate of only 16.8\% across both seen and unseen classes. The poor performance is likely attributable to the limited exposure of the pre-trained model to the specific prompt templates used in the VQA task, in contrast to the prompt templates frequently encountered by the pre-trained vision-language models in the first two tasks. Nevertheless, we observe that all prompt learning methods significantly enhance the performance of the pre-trained language model on the VQA task. Notably, in this more challenging vision task, the DiCap method continues to lead, improving the CLIP model's predictive performance by an average of 34.6\%.

\subsubsection{Counterfactual Quality Evaluation}
\label{ex:cf_ev}

We conduct experiments on the ImageNet dataset, adjusting the scale parameter \(s\) in Eq. \ref{eq:intervention} to control the quality of counterfactuals and analyze its impact on prompt learning performance. According to \citet{sanchez2022diffusion}, a smaller scale results in images nearly identical to the factual ones, while a larger scale causes the loss of important non-causal information. Therefore, both excessively large and small \(s\) affect the quality of counterfactuals. To quantify the impact of \(s\) on counterfactual quality, we adopt the CLD metric defined in \citet{sanchez2022diffusion} (\textbf{lower CLD indicates higher counterfactual quality}) and compare the CLD scores and predictive performance of the DiCap and CPL methods under optimal settings. 

The results, shown in Figure \ref{fig:scale}, reveal that lower CLD scores consistently correlate with improved prompt learning performance, highlighting that high-quality counterfactuals are critical for performance optimization. Furthermore, within a broad range (\(5 < s < 20\)), the DiCap method consistently achieves lower CLD scores than CPL, demonstrating its ability to generate higher-quality counterfactuals, which is the fundamental reason for its superior performance compared to CPL. 
 
\begin{table}[t]
    \centering
    \vskip 0.15in
    \renewcommand\arraystretch{1.2}
    \setlength{\tabcolsep}{3pt}
    \begin{tabular}{ccccccc}
    \hline
         \textbf{Prompt}&\multicolumn{2}{c}{\textbf{ImageNet}} &\multicolumn{2}{c}{\textbf{MSCOCO(1.5\%)}} &\multicolumn{2}{c}{\textbf{VQAv2(0.5\%)}}\\ 
         \cmidrule(lr){2-3} \cmidrule(lr){4-5} \cmidrule(lr){6-7}
        \textbf{Length }& \textbf{Seen}&\textbf{Unseen}& \textbf{Seen}&\textbf{Unseen}& \textbf{Seen}&\textbf{Unseen} \\ \hline
        
\textbf{4} &\textbf{86.07}	&\textbf{84.77}	&\textbf{63.55}	&\textbf{75.95}	&\textbf{44.50}	&\textbf{25.50} \\ \hline
\textbf{8}&85.38	&82.94	&62.74	&74.81	&44.08	&24.56 \\ \hline
\textbf{16}&84.98	&80.13	&61.92	&74.02	&43.23	&23.84 \\ \hline
    \end{tabular}
    \caption{Ablation Study of Prompt Length. }
    \label{tab:length}
\end{table}
\subsubsection{Ablation Study}

\textbf{Stability Analysis:} We further examine the impact of the sole hyperparameter \({\lambda}\) in Eq. \ref{eq:total_loss} on model performance. As shown in Figure \ref{fig:lambda}, we present results for unseen classes on the ImageNet and MSCOCO datasets (using 1\% of samples from the training set). It is evident that DiCap's predictive performance remains highly stable across both datasets despite variations in \(\lambda\), consistently outperforming the vanilla instance-conditional prompt learning method ($\lambda = 0$). This indicates that our approach is robust to changes of \(\lambda\), demonstrating strong stability.

\textbf{Few-shot Evaluation:} Increasing the number of shots in the training data generally enhances predictive accuracy, a trend corroborated by our experiments. As illustrated in Figure \ref{fig:shot}, we assess the performance of CoCoOp, CPL, and our DiCap method on unseen classes of ImageNet with shots set to 4, 8, and 16. The results reveal a steady improvement in DiCap’s accuracy as the number of shots increases, with our approach consistently surpassing the other two methods across all shot configurations.

\textbf{Prompt Length Analysis:} Table \ref{tab:length} reveals that shorter prompts consistently lead to better generalization across all three benchmarks, particularly under unseen settings. We hypothesize that longer prompts may introduce redundant or task-specific cues that the model can exploit spuriously, hindering its ability to generalize beyond the training distribution. In contrast, shorter prompts act as a form of regularization, encouraging the model to focus on invariant causal features.
\subsubsection{Sampling Strategy}
\label{ex:sample}
As shown in Table \ref{tab:sample}, we compare two negative label sampling methods: one based on classification probability similarity, as described in Section \ref{causal_prompt_construct}, and the other involving random sampling of negative labels. We evaluate prompt learning with counterfactual images generated using each sampling strategy across ImageNet, MSCOCO with 1.5\% of training samples, and VQAv2 with 0.5\% of training samples. The results demonstrate that learning with more challenging samples, specifically, similarity-based sampling, further enhances DiCap’s predictive accuracy on both seen and unseen classes, validating the effectiveness of our sampling strategy.

\begin{table}[t]
    \centering
    \vskip 0.15in
    \renewcommand\arraystretch{1.2}
    \setlength{\tabcolsep}{3pt}
    \begin{tabular}{ccccccc}
    \hline
         \textbf{Sampling}&\multicolumn{2}{c}{\textbf{ImageNet}} &\multicolumn{2}{c}{\textbf{MSCOCO(1.5\%)}} &\multicolumn{2}{c}{\textbf{VQAv2(0.5\%)}}\\ 
         \cmidrule(lr){2-3} \cmidrule(lr){4-5} \cmidrule(lr){6-7}
        \textbf{Strategy }& \textbf{Seen}&\textbf{Unseen}& \textbf{Seen}&\textbf{Unseen}& \textbf{Seen}&\textbf{Unseen} \\ \hline
        \textbf{Random	}&85.56	&84.08	&62.98	&74.22	&43.83	&23.13 \\ \hline
\textbf{Similarity} &\textbf{86.07}	&\textbf{84.77}	&\textbf{63.55}	&\textbf{75.95}	&\textbf{44.50}	&\textbf{25.50} \\ \hline
    \end{tabular}
    \caption{\textbf{Performance comparison} between two \textbf{sampling strategies} of $DiCap$ on three visual tasks. }
    \label{tab:sample}
\end{table}

\section{Conclusions}
\label{conclu}

This work delves into leveraging causality-driven diffusion models to refine prompt representations by aligning them with the causal characteristics inherent in images, thereby fostering the predictive performance of vision language models, notably on unseen classes. Our approach is underpinned by a robust theoretical foundation, providing stringent error bounds, an aspect not yet captured by existing literature. Extensive experimental results substantiate the robustness of our method across diverse visual tasks, highlighting the pivotal role of counterfactual generation in enhancing the performance of prompt learning. While our current model is primarily tailored for uni-modal prompt learning, it exhibits significant potential for seamless adaptation to multi-modal settings. We leave these avenues for future exploration. We also emphasize the importance of responsible deployment of generating counterfactuals, especially in sensitive domains such as healthcare.

\bibliographystyle{ACM-Reference-Format}
\bibliography{sample-base}

\end{sloppypar}
\end{document}